# Efficient selective attention LSTM for well log curve synthesis

Yuankai Zhou, Division of Geophysical and Geochemical Exploration, China National Nuclear Corp Beijing Research Institute of Uranium Geology, Beijing, China, ykzhou0824@gmail.com
Huanyu Li, College of Computer Science and Technology, Zhejiang University, Hangzhou, China, 22121003@zju.edu.cn

**Abstract**
Non-core drilling has gradually become the primary exploration method in geological exploration engineering, and well logging curves have increasingly gained importance as the main carriers of geological information. However, factors such as geological environment, logging equipment, borehole quality, and unexpected events can all impact the quality of well logging curves. Previous methods of re-logging or manual corrections have been associated with high costs and low efficiency. This paper proposes a machine learning method that utilizes existing data to predict missing data, and its effectiveness and feasibility have been validated through field experiments. The proposed method builds on the traditional Long Short-Term Memory (LSTM) neural network by incorporating a self-attention mechanism to analyze the sequential dependencies of the data. It selects the dominant computational results in the LSTM, reducing the computational complexity from $O(n^2)$ to $O(nlogn)$ and improving model efficiency. Experimental results demonstrate that the proposed method achieves higher accuracy compared to traditional curve synthesis methods based on Fully Connected Neural Networks (FCNN) and vanilla LSTM. This accurate, efficient, and cost-effective prediction method holds a practical value in engineering applications.

## 1. Introduction

Well logging is an essential technique in geophysical exploration, plays a significant role in lithology classification, formation evaluation, and reserve calculation (Cai et al.2015) (Czubek, 1972). With well log curves serving as the primary carriers of geological information, their importance has grown substantially in geological exploration research and production activities in recent years. However, objective factors such as geological conditions, measurement instruments, and drilling parameters can impact the quality of well logging data, resulting in outliers, consecutive gaps, or shifts in the data (Humphreys et al. 1983). These data quality issues directly influence the accuracy of well log interpretation, subsequently affecting model establishment and subsequent production development. In addressing these challenges, conventional approaches involve repeated well logging or predicting missing segments based on available information. Nonetheless, the high costs associated with repeated well logging necessitate avoidance in practical engineering production. Reconstruction methods can be categorized into interpolation-based methods and model-based synthesis methods. Interpolation-based methods are concerned with estimating values between missing data points (Gei et al. 2023) (Tang et al. 2004), while model-based methods focus on building a complete sequence model to synthesis missing data (Singh et al. 2021).

In this study, the neural network we used belongs to a model-based method. The development of neural networks in well logging curve synthesis has undergone two stages: simple artificial neural networks and recurrent neural networks (Hethcoat et al. 2019) (Zhu et al. 2022) (Ao et al. 2019). Simple neural networks, also known as multilayer perceptrons or Fully Connected neural networks. Bhatt and Helle (2002) utilized a multivariate regression-based neural network approach to uncover nonlinear features inherent in well log curves. This approach facilitated the prediction of permeability from well log data, yielding generally accurate trends but with some amplitude discrepancies. The overall predictive quality was moderate, necessitating substantial data preprocessing. Rolon et al. (2009) employed artificial neural networks to synthesize well log curves for analyzing reservoir characteristics in missing segments. Following a detailed analysis of the physical significance of various parameters, they generated γ, density, neutron, and resistivity well log data. Results indicated that the neural network model, compared to traditional methods like multivariate regression, exhibited distinct advantages in the accuracy of synthesized curves. However, the model's simplicity in structure and limited expressive power, coupled with its high demand for data quality, led to reduced performance in geologically complex areas. Zerrouki et al. (2014) synthesized data using an artificial neural network combined with clustering for lithology identification and reserve estimation. The results demonstrated the stability of nonlinear features extracted by artificial neural networks from well logging data. When recurrent neural networks emerged, people realized that the recurrent structure, which can capture contextual information from input sequences, is more suitable for handling curve synthesis tasks. Zhang et al. (2018) designed a neural network model centered around Long Short-Term Memory layers for well logging curve synthesis. By deepening the network layers, they achieved log synthesis, emphasizing inter-curve correlations and adjusting the network structure based on their own data. Their work compared several common synthesis methods, including Fully Connected Neural Networks and standard LSTM, ultimately demonstrating the superior performance of their proposed cascaded LSTM. In subsequent research, people focused on enhancing the feature extraction part using different methods. Zaremba et al. (2014) combined Convolutional Neural Networks (CNN) with LSTM, incorporating a convolutional layer before LSTM input. The convolutional layer extracted intrinsic correlations of current batch information, enhancing these features for LSTM training and reducing data dimensions for LSTM input.

To summarize the above research, the limitations of applying neural network methods in curve synthesis tasks are reflected in the following three aspects:

(1) Insufficient global modeling capabilities: Fully Connected networks cannot capture the contextual information of input sequences, while RNN-based LSTM networks exhibit poor performance in handling long sequences, which come with the risk of vanishing or exploding gradients.

(2) Inefficient training process: Due to its fixed computation framework, the LSTM exhibits low speed during training.

(3) Insufficient flexibility and interpretability: Both Fully Connected networks and LSTM networks have a fixed framework, meaning that we cannot directly discern the relationship between input parameters and outputs. Parameter adjustment is done blindly, requiring numerous experiments in neural networks, thereby increasing the cost of use.

In our model, we utilized the self-attention mechanism, which has the ability of global modeling to enhance a LSTM network. Therefore, the model can capture the contextual information of inputs, alleviating the problem of gradient anomalies. Meanwhile, we utilized a select layer to reduce the sequence length that will be sent into the LSTM layer, allowing only the positions with high attention weight to change the weight of the LSTM. This mechanism can reduce complexity and improve training efficiency. We name our model ESA-LSTM (Efficient Selective Attention LSTM). Finally, through analyzing the attention matrix, we can directly observe important parts of inputs. By adjusting the weight of different parts to enhance the training process, this transparent process allows us to clearly understand the model's judgment. Meanwhile, we can improve the model's interpretability.

## 2. Methodology

The synthesis of well logging curves is a typical many-to-one prediction problem, well logging data consists of multiple parameters. To predict the missing data using existed data, we need to establish correspondences between the parameters. The mathematical model for this problem can be described by Formula 1:

$$\hat{y} = f_w(x_1, x_2, \ldots, x_n) + b \tag{1}$$

In Formula 1, $\hat{y}$ represents the variable to be predicted, $x$ denotes the available parameters, $f$ represents the nonlinear mapping relationship between the parameters, which is the model, $w$ represents the parameters within the model, and $b$ is the bias term. In a neural network model, by selecting appropriate activation functions, network structures, training, and optimization strategies, a nonlinear mapping relationship between the input parameters and the output can be constructed. During the training process, the network automatically adjusts the parameters and biases through iterations to minimize the differences between predicted values and actual values.

The framework of the proposed approach is illustrated in Figure 1. This method comprises three main components. Firstly, the foundational LSTM framework, constructed with LSTM layers. Secondly, our select layer, which consists of a self-attention layer and a Top-K selector. Lastly, the input and output layers, each accompanied by a fully connected layer positioned after the input layer and before the output layer.

### 2.1 Long Short-Term Memory Network

The LSTM used in this paper is based on Recurrent Neural Networks (Yu et al. 2019), which is a type of machine learning model as illustrated in Figure 2 (a). Due to its recursive structure, RNN can retain partial information from previous input data, making them widely used for handling sequential data. In the well logging curve synthesis problem, well logging data exhibits significant correlations in continuous geological layers.

For an input sequence $X(x_1, x_2, \ldots, x_t)$, where $x_t$ represents the input at time step $t$, RNN uses a hidden state $H(h_1, h_2, \ldots, h_t)$ to retain the "historical information" of the input. The hidden state can be expressed as Formula 2:

$$h_t = \sigma(w_{xh}x_t + w_{hh}h_{t-1} + b_h) \tag{2}$$

In Formula 2, $w_{xh}$ is the weight matrix between the input layer and the hidden layer, $w_{hh}$ is the weight matrix between the hidden layer and the next hidden layer, and $\sigma$ denotes the activation function. $b_h$ represents the bias value. By outputting the hidden state at a certain time step, $y_i$ can be expressed as Formula 3:

$$y_t = g(W_{hy}h_t + b_y) \tag{3}$$

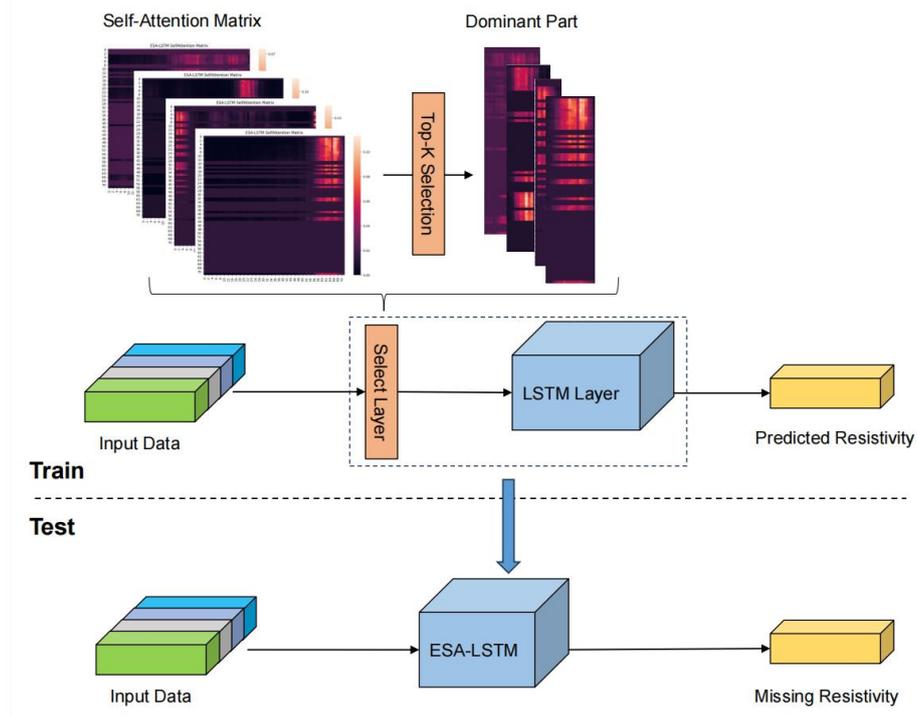

**Fig.1** The schematic diagram of the approach in this paper. In this figure, we take predicting resistivity as an example. During the training phase, the "Select Layer" is used to compute the self-attention matrix for the current input batch. From this matrix, K dominant computations are selected, which are the only inputs sent to the LSTM layer. The output of the LSTM layer is then combined with the non-selected part and fed into the fully connected layer in the backend, which produces the final prediction result. During the testing phase, existing data is fed into the trained network to infer the missing data.

In Formula 3, $W_{hy}$ is the weight matrix between the hidden layer and the output layer, g denotes the activation function of output layer, $b_y$ represents the bias value.

To address the issue of uncontrolled gradients in the recursive process of RNN, the concept of Long Short-Term Memory (Hochreiter and Schmidhuber. 1997) (Graves. 2012) has been integrated into the general RNN network. The core idea of LSTM lies in the design of a set of gate units that control the input data through an input gate, reduce values towards zero through a forget gate, and use the hidden state through an output gate. The structure of the LSTM network is illustrated in Figure 2 (b), where a new variable $C(c_1, c_2, \ldots, c_t)$, referred to as the cell state, is introduced. The value of the cell state depends on the values of the three gate units. Therefore, the model can be updated as Formula 4:

$$\begin{aligned} i_t &= \sigma(W_i(h_{t-1} + x_t) + b_i) \\ f_t &= \sigma(W_f(h_{t-1} + x_t) + b_f) \\ c_t &= f_t \cdot c_{t-1} + i_t \cdot \tanh(W_c \cdot (h_{t-1} + x_t) + b_c) \\ o_t &= \sigma(W_o \cdot (h_{t-1} + x_t) + b_o) \\ h_t &= o_t \cdot \tanh(c_t) \end{aligned} \quad (4)$$

In Formula 4, $i_t$, $f_t$, and $o_t$ represent the states of the input gate, forget gate, and output gate, respectively. $c_t$ is the alternative cell state. $h_{t-1}$ is the hidden state. $b_i, b_f, b_o$, and $b_c$ are the biases of the gate units, and $\sigma$ represents the activation function. During training, $o_t$ determines which part of the cell state will be pushed out as the hidden state. $h_t$ is a combination of the output gate and the cell state, while W representing the weights for each gate unit.

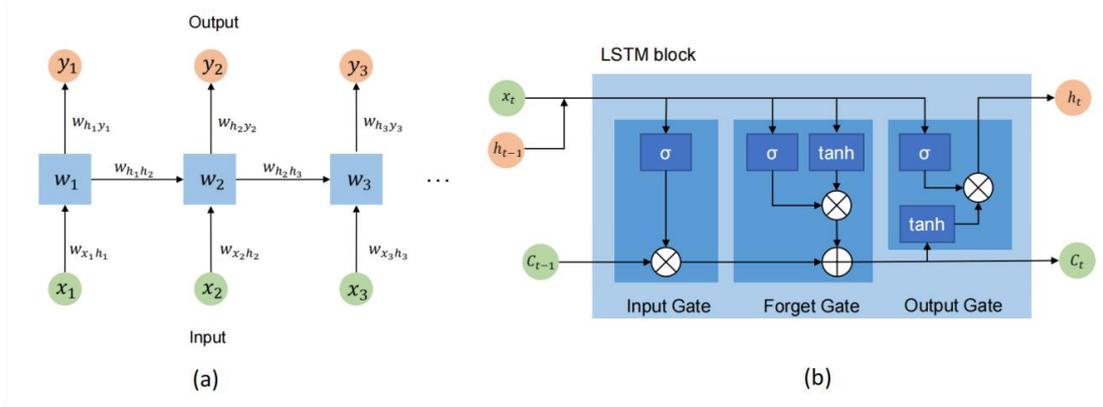

Fig.2 Schematic diagram of RNN (Figure a) and LSTM unit (Figure b). The green part in the figure represents the input, the blue part represents the network weight, and the red part represents the output hidden state. In (b), σ represents the activation function and tanh represents the hyperbolic tangent.

However, the window of context information that LSTM can capture is limited, while well logging data contain complex geological information. The local perspective of LSTM can restrict its ability. Therefore, we need to capture a broader range of spatial dependencies and more comprehensive context information.

**2.2 Self-Attention Mechanism**

Transformer is a deep neural network architecture proposed in 2017(Vaswani et al. 2017). The attention mechanism is the core encoding method of Transformer, and the calculation process is shown in Figure 3.

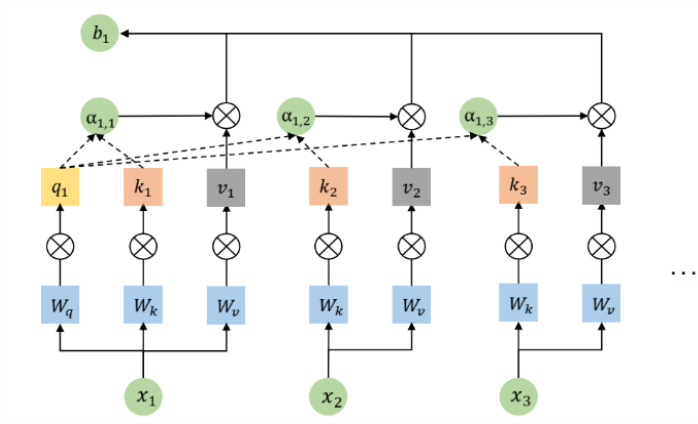

Fig.3 Schematic diagram of the calculation process of the attention matrix. In this figure, $x$ is the input data, and $W_q$, $W_k$, and $W_v$ are all learnable matrices, α is the cell of the self-attention matrix.

First, for an input $X(x_1, x_2, ..., x_t)$, we calculate the corresponding $Q(query) = X \cdot w_q$, $K(key) = X \cdot w_k$, and $V(value) = X \cdot w_v$, where $w_q$, $w_k$, and $w_v$ are learnable weight matrices, they will be updated as the model iterates. The second step is to compute the self-attention weights:

$$\alpha(i,j) = softmax\left(\frac{QK^T}{\sqrt{d_k}}\right)V \tag{5}$$

In Formula 5, $softmax$ is a function used to convert a set of values into a probability distribution. It is often used in transformers to calculate attention weights. Its mathematical expression is $softmax(x)_i = \frac{e^{x_i}}{\Sigma_j e^{x_j}}$. $d$ is the dimensionality of the current vectors. The attention weight corresponding to each moment is $b_i = \Sigma(\alpha(i, j) \cdot v_j)$. To express more features, multiple sets of $Q$, $K$, and $V$ are used with corresponding weight matrices, and their results are concatenated and transformed, forming the multi-head self-attention mechanism. In addition, this process is calculated in parallel, so a position encoding needs to be added at the end to restore the position information.

The self-attention mechanism enables the model to obtain global contextual information, further improving the model's performance when processing long sequences. However, although self-attention is calculated in parallel, the complete Transformer encoding and decoding process will still contain a large number of matrix operations, which will increase the number of parameters of the model. Therefore, our method only calculates the self-attention matrix, filters the input data based on the results of the self-attention matrix, and selects the parts that dominate the results. The backbone of the model is still LSTM, so that the extracted features contain sufficient context. information, and avoids too many complex calculations, while also reducing the impact of low-correlation inputs on model convergence. The method framework of this article is shown in Figure 5.

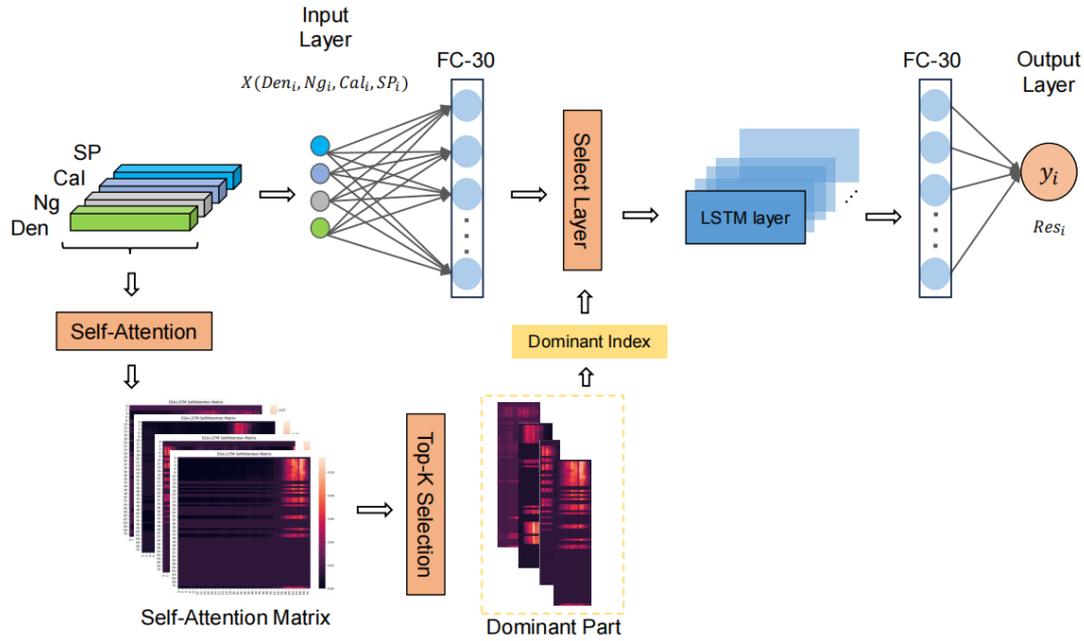

**Fig.5** Schematic diagram of ESA-LSTM, which demonstrates the resistivity prediction process of a point. This process will continue in a loop, and the selection layer will determine whether to open the input gate for the current input based on the incoming index each time. After the current batch input is completed, the location information is restored using location encoding.

**3.Experiments and Results**

This article uses uranium mine logging data from the Nalinggou area in the northeastern Ordos Basin. The caprock in the northern Ordos Basin includes the Mesozoic and Cenozoic Triassic ($T$), Jurassic ($J$),

Lower Cretaceous ($K_1$), Paleogene Oligocene ($E_3$), and Neogene Upper Pliocene ($N_2$), Quaternary ($Q$), among which the Triassic, Jurassic and Cretaceous are the main sedimentary cover of the basin (Wang et al. 2015) (Zhu et al. 2019). The lithology of the Nalinggou area is medium sandstone and coarse sandstone, and the colors are mainly gray, gray-green, green, etc. The sandstone has a low degree of consolidation, frequent sand-mud interbeds, and multiple positive rhythmic deposits developed from top to bottom. Spin. The logging parameters used in the experiment include resistivity, density, natural gamma, well diameter, natural potential, and other data. The sampling interval is 0.05m. We selected data from 50m to 350m deep for each well, a total of 20 wells, of which 15 were used for training and 5 were used for testing. In the result display, the 5 wells used for testing are marked A~E.

**3.1 Setup**

The experimental section of this paper consists of two parts: curve synthesis and ablation experiments. In the curve synthesis task, we utilized four parameters: resistivity, density, natural gamma, and caliper. For each parameter, we intentionally set one of them as a missing value and employed the other three parameters to predict it. Additionally, we recorded the average training time for each model. We compare our proposed ESA-LSTM with Fully Connected Neural Networks (FCNN), vanilla LSTM, and the cascaded LSTM (C-LSTM). The vanilla LSTM has a 30-dimensional fully connected layer in the input and output layers, where the input layer maps the parameter characteristics to 30 dimensions. Thus, the LSTM layer also has a dimensionality of 30. The output layer integrates the results from the LSTM layer into a single output, representing the predicted value of the missing curve at the current depth, which is compared with the ground truth. The C-LSTM increases the number of layers of vanilla LSTM, which has been shown to enhance the model's expressive power and capture a wider range of long-term dependencies. However, increasing the number of layers in this way also increases the computational cost and introduces potential gradient issues due to the deep network. Our improvement on the basic structure is to add a self-attention mechanism and a select layer after the first fully connected layer. The aim is to select the portion with the highest average self-attention weights from the current batch and feed those values into the LSTM layer. The remaining parts no longer have the ability to change the hidden layer parameters when passing through LSTM. Finally, we use position encoding to restore the position information of the two parts. The root mean square error (RMSE) is used as the evaluation metric in the experiments mentioned above.

To validate the effectiveness of the self-attention mechanism we employed, we visualized the self-attention matrix using a batch of data. The results are shown in Figure 6, where brighter regions indicate higher relevance. These regions have a more dominant influence on the weight changes in the LSTM. Furthermore, we utilized an 8-headed self-attention mechanism, it uses 8 sets of feature matrices to analyze the intrinsic correlation of signals. In the figure 6, it can be observed that in actual well logging data, only a few positions have higher values in the self-attention matrix, while the remaining lower values introduce significant redundancy in computations. By preventing these lower-weighted regions from entering the LSTM layer, the convergence of the model can be accelerated with minimal impact on the model's accuracy.

Figure 7 depicts the presentation of all prediction results, with the blue solid line representing the original data and dashed lines indicating predictions from various models. Figure 8 provides an in-depth showcase of the resistivity parameter alone, where the blue denotes the original data, and the orange represents the model's predicted outcomes. Tables 1 and 2 correspondingly display the RMSE results. Figure 9 displays the results of synthesizing resistivity for multiple wells using the same model. Correspondingly, Table 3 presents its RMSE results. To analyze different components of the model, we conducted two sets of ablation experiments: one focusing on the number of neurons in the fully connected layer, and another experiment focuses on the proportion of features entering LSTM. To investigate the influence of the number of neurons in the fully connected layer on model performance, we set four options for comparison: 10, 30, 50, and 100. The results are shown in Table 4. In addition, we use a select layer to reduce the amount of data input to the LSTM, to verify the effectiveness of the selection layer, we select 0 (no selection), 10%, 30%, 50% and 100% (all selections) into LSTM. The synthesis results are shown in Table 5.

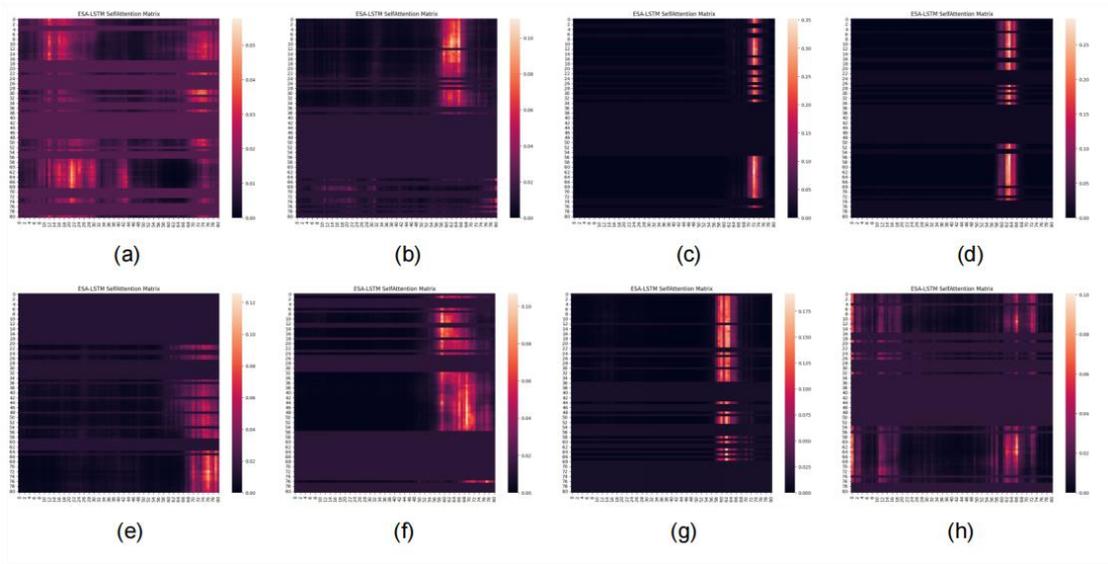

**Fig.6** Visualization of the Attention Matrix, where the brighter regions in the figure represent the high values in the self-attention matrix. These regions have a significant impact on the generated output, indicating strong correlations.

### 3.2 Results

**Table.1** Results of synthesizing different parameters for the same well using different models, along with the average computation time.

| Model | Res | Den | Ng | Cal | Average time |
|---|---|---|---|---|---|
| FCNN | 1.312 | 0.043 | 5.509 | 4.053 | 11" |
| Cascaded LSTM | 0.617 | 0.021 | 2.561 | 3.539 | 1'53" |
| ESA-LSTM | 0.424 | 0.004 | 0.635 | 0.348 | 42" |

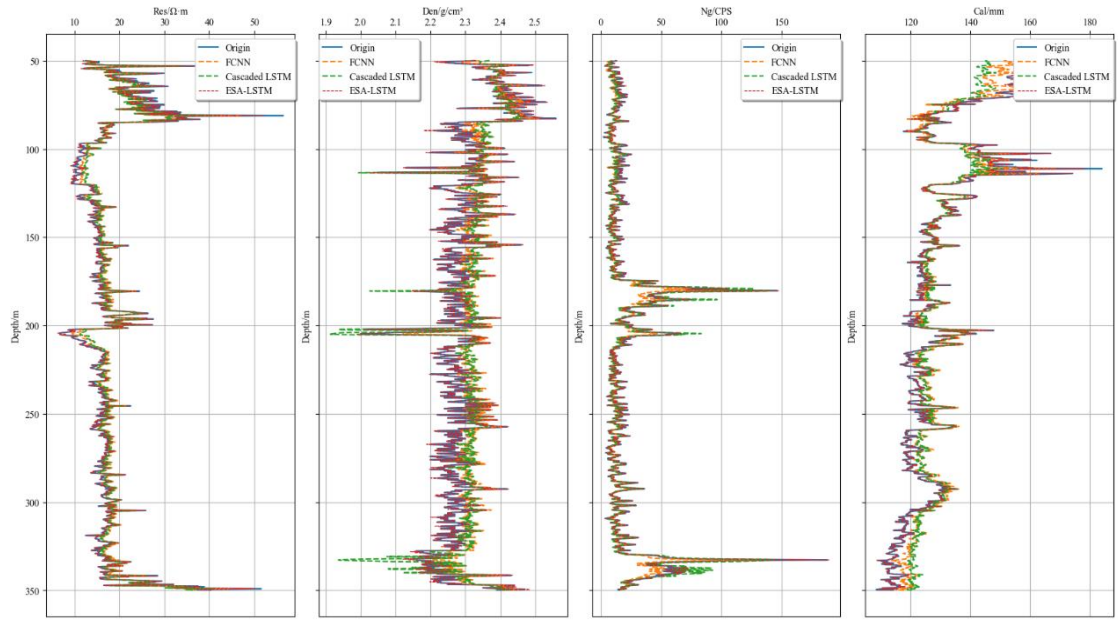

**Fig.7** Comparison of results for parameter synthesis using different models within the same well. From left to right: resistivity, density, natural gamma, and caliper; the blue solid line represents actual values, while the dashed lines depict predictions from different models.

**Table.2** RMSE values of resistivity for different models

| Model | RMSE |
|---|---|
| ESA-LSTM (Ours) | 0.424 |
| LSTM | 0.974 |
| Cascaded-LSTM | 0.617 |
| FCNN-4 | 1.312 |
| FCNN-8 | 1.306 |

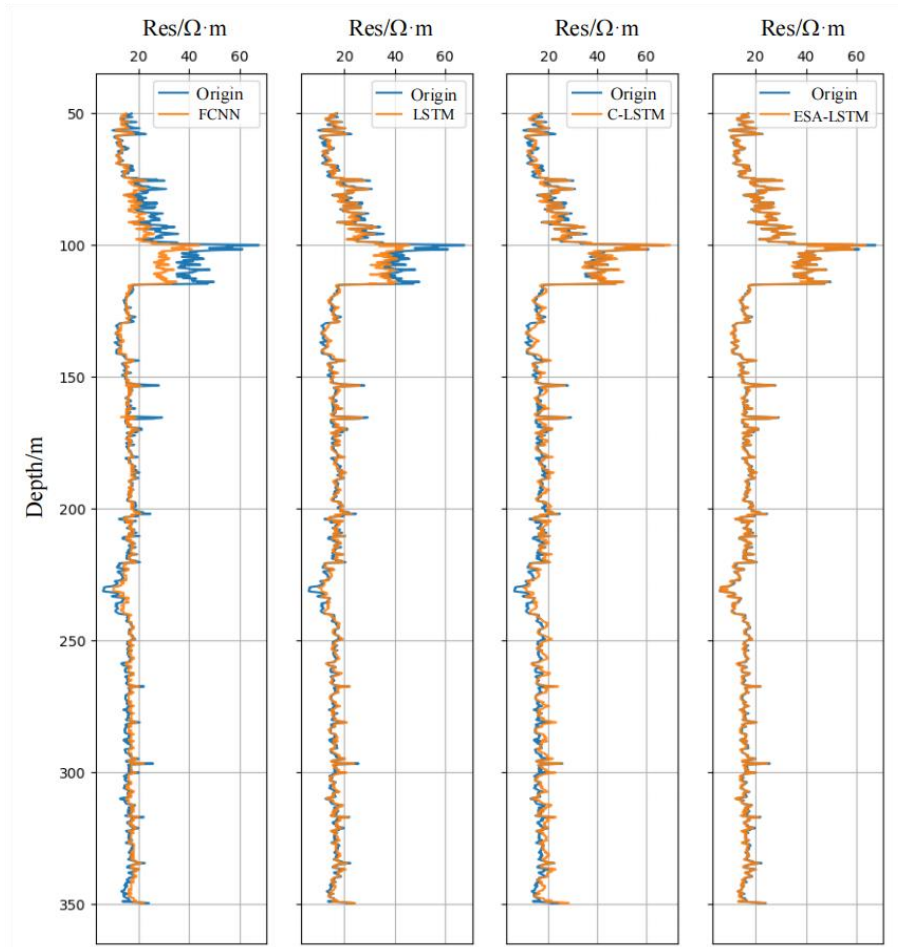

**Fig.8** The results of the curve synthesis, from left to right are the comparison results of the original data and the fully connected layer, vanilla LSTM, Cascaded LSTM, and our ESA-LSTM.

**Table.3** Results of resistivity synthesis for different wells using the same model.

| Model | A | B | C | D | E |
|---|---|---|---|---|---|
| FCNN | 3.247 | 2.164 | 1.312 | 0.981 | 1.800 |
| Cascaded LSTM | 2.559 | 1.788 | 0.617 | 0.899 | 1.534 |
| ESA-LSTM | 0.525 | 0.598 | 0.324 | 0.298 | 0.327 |

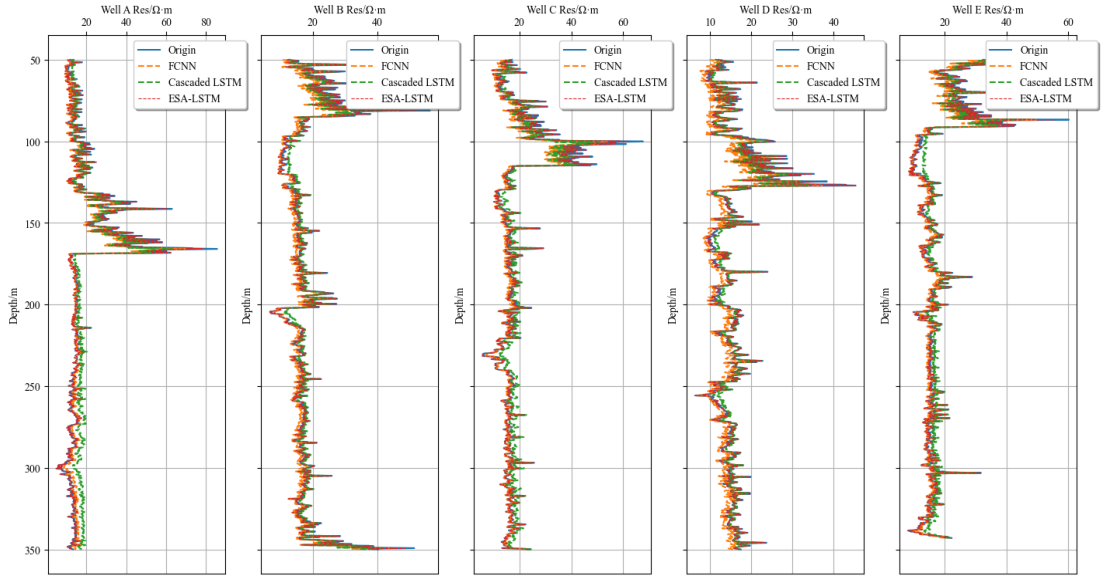

**Fig.9** Synthesis of resistivity results for multiple wells using the same model. From left to right: Well A to Well E.

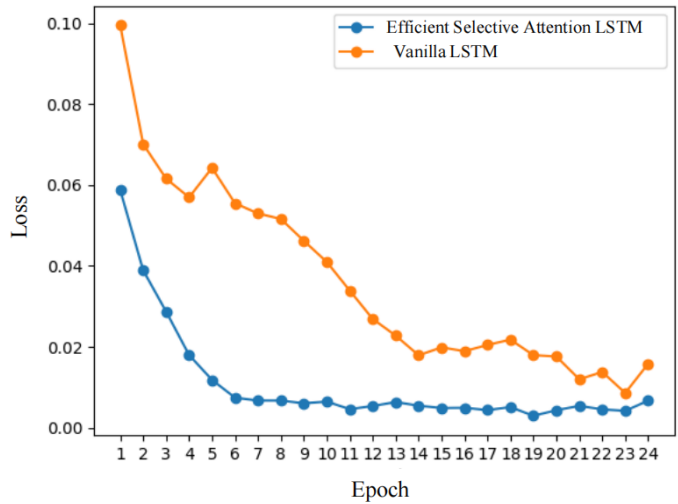

**Fig.10** Training history, the blue line in the figure represents ESA-LSTM, and the orange line represents the basic LSTM model.

**Table.4** Synthesis results corresponding to different numbers of neurons in the fully connected layer, the first column in the table is the number of the well.

| Well | RMSE for varying neuron counts in FC layer | | | |
| :---: | :---: | :---: | :---: | :---: |
|  | 10 | 30 | 50 | 100 |
| A | 0.510 | 0.388 | 0.870 | 2.417 |
| B | 0.517 | 0.241 | 0.339 | 1.661 |
| C | 0.521 | 0.245 | 0.561 | 2.063 |
| D | 0.522 | 0.184 | 0.296 | 1.217 |
| E | 0.448 | 0.217 | 0.365 | 1.511 |

**Table.5** Select the synthesis results corresponding to the high values of the self-attention matrix with different percentages, where 0 means no selection, that is, skip the LSTM layer and directly enter the fully connected layer, and 100% corresponds to all entering the LSTM layer, which is the basic LSTM network.

| Well | RMSE for different self-attention percentage options | | | | |
|---|---|---|---|---|---|
| | 0 | 10% | 30% | 50% | 100% |
| A | 1.452 | 0.653 | 0.458 | 0.424 | 0.974 |
| B | 1.112 | 0.457 | 0.309 | 0.204 | 0.886 |
| C | 1.214 | 0.608 | 0.479 | 0.374 | 0.868 |
| D | 1.151 | 0.360 | 0.276 | 0.209 | 0.714 |
| E | 0.988 | 0.418 | 0.268 | 0.255 | 0.911 |

## 4. Discussion

In this study we ultilized the attention mechanism to strengthen the LSTM network, conducts synthetic log curve generation experiments, and verifies the impact of the number of neurons and the proportion of our selection layer selected into the LSTM on the results. In the synthetic log experiment, we first assume that a well has lost resistivity, density, natural gamma and well diameter data in sequence. The above logging parameters were synthesized separately using ESA-LSTM trained on the data of 15 wells; subsequently, we used the model to synthesize the resistivity curves of the five test wells. ESA-LSTM achieved the smallest RMSE in both parts of the experiment, and showed higher efficiency and faster convergence speed. Our ESA-LSTM can effectively extract patterns with long-term spatial dependence and estimate and reconstruct well logs based on these patterns. Furthermore, we compare Fully Connected Neural Networks, vanlla LSTM and cascaded LSTM with our ESA-LSTM. ESA-LSTM is better than the above models when dealing with well log curve synthesis problems.

In the first phase of ablation experiments, the impact of different numbers of neurons within the fully connected layer on network performance was investigated. Neurons serve as units to approximate the functional relationships between existing parameters and the target parameters to be predicted. In theory, employing an appropriate number of neurons allows for the fitting of any functional relationship; however, excessive neurons result in an increased parameter count, escalating computational costs and thereby diminishing model efficiency. As evidenced by Table 3, an increasing number of neurons leads to a decreasing then increasing trend in Root Mean Square Error (RMSE). When the number of neurons is at or below 10, the model is characterized by simplicity and inadequate expression capacity. In contrast, an increased number of neurons results in significant overfitting, degrading performance on unseen testing sets and reducing generalization capacity. Consequently, in the context of our well log curve synthesis problem, the most optimal number of neurons within the fully connected layer is determined to be 30.

In the second phase of ablation experiments, an analysis was conducted on the influence of different percentages of inputs entering the LSTM layer on model performance. Experimental outcomes suggest a trend in RMSE reduction followed by an increment as different percentages are selected. Early-stage reduction can be attributed to the model's proximity to a fully connected neural network with only two layers, which results in simplicity and limited expression capacity. As the selection of entries increases, low values within the self-attention matrix are inevitably introduced. This is because as the selection ratio increases, the model approaches a normal LSTM. According to the results, for the well log curve synthesis task of this article, the optimal selection range is around 30%. This ratio is subject to variations based on specific geological information; regions with complex geological structures necessitate more intricate models for fitting geological characteristics, resulting in a higher proportion of contributions from the primary computational section. Additionally, an increase in parameters such as the number of training epochs, the number of neurons in the fully connected layer, and the number of LSTM layers does not directly enhance model performance. On the contrary, overly complex models and over-learning will lead to overfitting of the model, thereby reducing the generalization ability. Moreover, it is observed that the accuracy of the model is relatively low when synthesizing certain parameter curves (e.g., natural gamma). This phenomenon shows that our model's synthetic performance of different parameters is not balanced, and the model does not fully capture the characteristics of the natural gamma curve. In future research, we may improve the data quality by changing the model structure or using anomaly detection methods. To further improve model performance and efficiency, more suitable architectures can be integrated on this foundation to optimize feature representation. It is our aspiration to elevate the performance and reliability of neural network models in well logging curves synthesis tasks. The results reflected in Table 4 indicate favorable outcomes when selecting 30% and 50% of the self-attention matrix entries. Furthermore, Figure 10 shows how the error changes with training epochs, illustrating the capacity of this approach to expedite convergence and enhance training efficiency.

## 5. Conclusion

The proposed ESA-LSTM model, based on LSTM architecture, effectively addresses the tasks of data completion and data quality enhancement in well logging tasks, thereby reducing the associated costs and workload. The model exhibits significant superiority in the domain of well log curve synthesis. Leveraging the LSTM structure and self-attention mechanism, the model adeptly captures inter-data correlations, thereby enhancing the accuracy of synthesized curves. The ESA-LSTM constructed in this article uses the attention mechanism to strengthen the basic LSTM structure, strengthen the network's ability to capture contextual information, and alleviate the risk of gradient anomalies. At the same time, the selection layer in this article reduces the amount of data input to the LSTM model, thereby improving the training efficiency of the model while maintaining accuracy. Finally, the decision-making mechanism of our model selection layer is transparent, and we verify the impact of different components on the synthesis results through ablation experiments. These enhance the interpretability of our work. For different geological environments, the hyperparameters in the model can be directly changed, which eliminates the limitations of the previous fixed models. This approach boasts minimal data requirements, rapid convergence rates, and robust versatility, rendering it valuable for practical engineering applications. Although the model showed high accuracy and generalization in the current area, when we transferred it to drilling in other areas, the performance dropped significantly, which is related to the regularization method of the parameters. Furthermore, the approach is now purely data-driven, with extensive work demonstrating that knowledge-driven approaches can increase model interpretability while improving model accuracy. At the same time, in order to be more widely used in actual projects, we will also work on simplifying the use process of this method through packaging software.

**Competing Interests**

We declare that we have no financial and personal relationships with other people or organizations that inappropriately influence our work, there is no professional or other personal interest of any nature or kind in any product, service and/or company that could be construed as influencing the position presented in, or the review of, the manuscript entitled.